\documentclass[letterpaper,10pt,conference]{ieeeconf}
\IEEEoverridecommandlockouts
\usepackage[T1]{fontenc}
\usepackage{amsmath,amssymb}
\usepackage{graphicx,booktabs,array,tabularx}
\usepackage{url}
\usepackage{xcolor}
\usepackage{tikz}
\usepackage{float}
\usetikzlibrary{arrows.meta,positioning,calc,fit,backgrounds}
\usepackage{microtype}
\usepackage{balance}

\newif\ifanonymous
\anonymousfalse
\newcommand{\formed}{\textsc{Formed}}
\newcommand{\crooked}{\textsc{Crooked}}
\newcommand{\emptygrasp}{\textsc{Empty}}
\newcommand{\state}[1]{\textsc{#1}}

\newcolumntype{Y}{>{\raggedright\arraybackslash}X}
\definecolor{bodyblue}{RGB}{34,79,121}
\definecolor{contactteal}{RGB}{19,112,110}
\definecolor{modeorange}{RGB}{151,85,24}

\title{\LARGE\bf
AgenticTactileVLA: Contact-Guided Execution-Time Supervision \\
for Generalizable Dexterous Manipulation without VLA Retraining
}

\ifanonymous
\author{Anonymous authors}
\else

\author{
Elizaveta Semenyakina$^{1,2,*}$,
Ivan Snegirev$^{1,*}$,
Mikhail Kiselev$^{1,*}$,
Miguel Altamirano Cabrera$^{1,2}$,\\
Artem Lykov$^{1,2}$,
Hajira Amjad$^{1}$,
and Dzmitry Tsetserukou$^{1}$
\thanks{$^{1}$Intelligent Space Robotics Lab, Skolkovo Institute of Science
and Technology, Moscow, Russian Federation.}
\thanks{$^{2}$R\&D Center, MWS, Moscow, Russian Federation.}
\thanks{$^{*}$These authors contributed equally to this work.}
\thanks{E-mails:
\{Elizaveta.Semenyakina, Ivan.Snegirev, Mikhail.Kiselev,
m.altamirano, Artem.Lykov, Hajira.Amjad,
D.Tsetserukou\}@skoltech.ru.}
}
\fi

\begin{document}

\bstctlcite{IEEEexample:BSTcontrol}

\maketitle
\thispagestyle{empty}
\pagestyle{empty}

\begin{abstract}
Vision-language-action policies may predict a transferable manipulation
strategy yet fail to realize it reliably on the encountered object: objects
compatible with the same grasp differ in geometry and compliance, and visual
feedback degrades under closure occlusion. AgenticTactileVLA is presented as an
execution-time supervisor that shifts part of object-specific adaptation from
prediction to physical interaction. A fixed VLA provides the approach and hand
targets; the supervisor decides whether to remain transparent, refine finger
flexion, retain or release the corrected configuration, return control to the
VLA for retry, or select a compliant hand-control regime. It uses
finger-position and motor-effort feedback as proprioceptive contact evidence
and requires neither tactile sensors nor VLA retraining. On a Unitree G1 with
a BrainCo Revo2 hand, a randomized matched-block evaluation on five objects
held out from VLA training yields 61.3\% completion for the base VLA, 72.0\%
for unconditional close-to-stall control, and 84.0\% for the supervisor under
a shared budget; the gain is positive on every object and persists under
moderate pose perturbations. Ablations show the gain is not explained by
extended closure alone, and that selective triggering reduces correction
episodes by 65.7\% with no detected change in completion. A retention audit
shows acceptance predicts retention in 88.9\% of held-out cases, while
compliant objects expose conservative false rejection. A thin-walled-cup
study demonstrates contextual routing to compliant control, matching an
always-compliant reference. These results suggest that contact-guided execution-time adaptation can improve the object-level generalization of a fixed VLA to held-out objects by adapting physical realization without object-specific retraining.
\end{abstract}

\section{Introduction}

A central challenge in robot learning is to generalize manipulation policies
to new objects while maintaining reliable physical execution. For
vision-language-action (VLA) models, this is particularly important in
contact-rich manipulation. Objects compatible with the same broad grasp
strategy may differ in geometry and compliance, changing the finger contacts
and interaction conditions required for successful execution. A VLA may
therefore generate a suitable approach yet still leave fingers without
sufficient support, produce an unsuitable terminal hand configuration, or
over-compress a fragile object. Such failures are difficult to resolve visually
during closure, when the hand increasingly occludes the contact region.

Many objects can be handled using a limited repertoire of broad grasp
strategies \cite{cutkosky1989grasp,feixgrasp}, while their physical realization
may still require object-specific finger contacts and interaction effort. This
motivates reusing a fixed policy while adapting these physical details during
execution rather than through object-specific retraining.

\begin{figure}[t]
\centering
\includegraphics[width=\columnwidth]{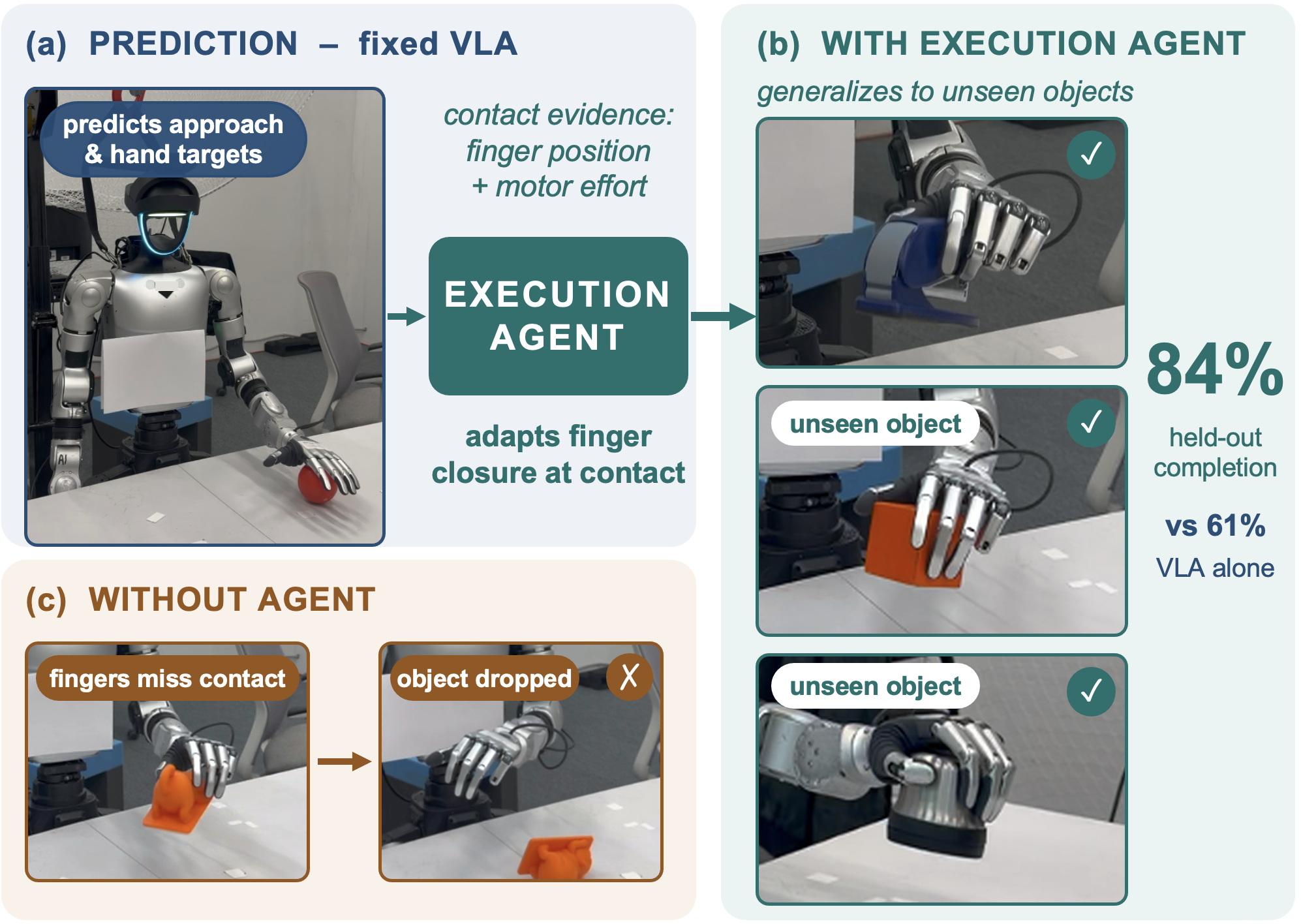}
\caption{From VLA prediction to execution-time physical adaptation.
The fixed VLA provides the approach and initial hand configuration, while the
execution agent adapts finger closure to the encountered object.}
\label{fig:platform}
\vspace{-4mm}
\end{figure}

We investigate a division of responsibility in which the VLA provides the
transferable manipulation strategy, while an execution-time supervisor adapts
its physical realization to the encountered object
(Fig.~\ref{fig:platform}). The supervisor can modify finger flexion and gate
progression of the VLA-generated body reference, but does not replace the VLA
approach policy or learn a residual action model. Prior work has explored
tactile-conditioned VLA adaptation \cite{vlatouch,unitacvla} and execution
monitoring and recovery \cite{planar,volo}. Our work lies at their intersection
and asks: \emph{Can execution-time physical adaptation improve the realization
of a fixed VLA policy on objects outside its training set without
object-specific retraining?}

AgenticTactileVLA augments a fixed VLA with a stateful execution supervisor.
At the terminal phase of nominal closure, it evaluates proprioceptive support
evidence. If support is sufficient, execution remains transparent to the
policy. Otherwise, body-reference progression is paused while whole-body
stabilization remains active, and unsupported fingers are corrected
independently. An accepted correction is retained; a rejected correction
releases the hand override and returns control to the VLA for subsequent
policy-generated retry.

A key principle is \emph{selective intervention}: the supervisor acts only when
the policy-proposed closure is not sufficiently realized physically. To
separate this effect from simply closing farther or longer, we compare against
unconditional close-to-stall control under the same execution budget and use
component ablations to examine selectivity, effort-assisted stopping, and
verdict-dependent release.

Geometric correction alone is insufficient when the interaction regime must
also change. Fragile objects may deform before proprioceptive resistance is
large enough for reactive correction. AgenticTactileVLA therefore also selects
between nominal position control and a predefined compliant mode before
closure. Explicit handling cues can request compliant interaction, while a
vision-language model (VLM) can select between the two modes from image and
task context. The VLM selects only the interaction mode and does not generate
controller parameters or material estimates.

We use the term \emph{agentic} to denote this bounded, stateful supervision:
deciding whether to remain transparent or intervene, adapting individual
finger references, accepting or releasing the corrected grasp, returning
control to the VLA for retry, and selecting the interaction regime. This distinguishes the layer from low-level reflexes by enabling high-level decision-making based on multimodal context. The
framework uses finger position and motor-effort signals as device-specific
proprioceptive evidence and requires neither additional tactile sensors nor
object-specific VLA retraining.

The main contributions of this work are:
\begin{enumerate}

\item A contact-guided execution-time supervisor for a fixed VLA, combining
selective intervention, bounded per-finger correction, grasp acceptance or
release, and policy-mediated retry.

\item A proprioceptive execution mechanism using finger position and
motor-effort feedback without additional tactile hardware, together with
language- or vision-conditioned selection between nominal and compliant
interaction modes.

\item Real-hardware evaluation on a Unitree G1 with a BrainCo Revo2 hand under
matched execution budgets, including unconditional close-to-stall control,
component ablations, pose perturbations, retention diagnostics, and
fragile-object interaction tests.

\end{enumerate}

The intended scope is adaptation of feasible policy-generated manipulation
strategies, not recovery from arbitrary incompatible approach errors.
Within this scope, we view execution-time physical adaptation as a step toward
more scalable object-level generalization of fixed VLA policies, reducing the
need for object-specific retraining while preserving the policy's reusable
manipulation strategy.

\section{Related Work}

\subsection{Contact Sensing from Motor Feedback}

Motor feedback can reveal physical robot--object interaction without dedicated
tactile sensors. Current as Touch \cite{currentastouch}, Transferring Contact,
Not Just Motion \cite{contacttransfer}, and Pseudo-Tactile
\cite{pseudotactile} use actuator or gripper signals for contact-aware
manipulation.

AgenticTactileVLA uses finger positions and motor-effort signals as
device-specific evidence of interaction resistance rather than calibrated
contact force. This evidence determines whether the nominal closure provides
sufficient support and, when necessary, guides local finger correction and
grasp acceptance or release.

\subsection{Tactile Feedback for VLA Adaptation}

Tactile feedback complements vision when physical interaction is difficult to
infer from images alone. VLA-Touch \cite{vlatouch} combines a frozen VLA with
semantic tactile feedback and learned action refinement. TLA \cite{tla} and
VTLA \cite{vtla} incorporate tactile observations into action generation,
while UniTacVLA \cite{unitacvla} predicts contact changes to refine action
chunks. RoboTacDex \cite{robotacdex} provides multimodal data for
visual--tactile dexterous humanoid manipulation.

In contrast, AgenticTactileVLA does not train a tactile-conditioned VLA or a
learned residual policy. Adaptation is placed in an external, nonlearned
execution layer while the VLA remains fixed. The supervisor stays transparent
when the policy-generated closure already provides sufficient support and
intervenes only when physical correction is required.

\subsection{Execution Supervision and Interaction Control}

Execution supervision connects predicted actions with observed consequences.
PLanAR \cite{planar} verifies expected effects and replans after deviations,
VoLo \cite{volo} combines monitoring with recovery, and VLAs Fail Differently
\cite{failsdifferently} analyzes failure structure in motor-command sequences.

This work brings supervision to the closure-to-motion transition in dexterous
manipulation. The supervisor determines whether the physical realization of
the VLA action is sufficient, whether individual fingers should be corrected,
whether the resulting grasp should be retained or released, and whether
control should return to the VLA for another approach. A separate
interaction-mode layer selects between nominal position control and
current-limited compliant control before closure.

AgenticTactileVLA thus combines selective intervention, contact-guided finger
adaptation, grasp acceptance or release, policy-mediated retry, and contextual
interaction-mode selection within a single execution-time layer over a fixed
VLA policy.

\section{Method}
\label{sec:method}

\begin{figure*}[!t]
\centering
\includegraphics[width=0.85\textwidth]{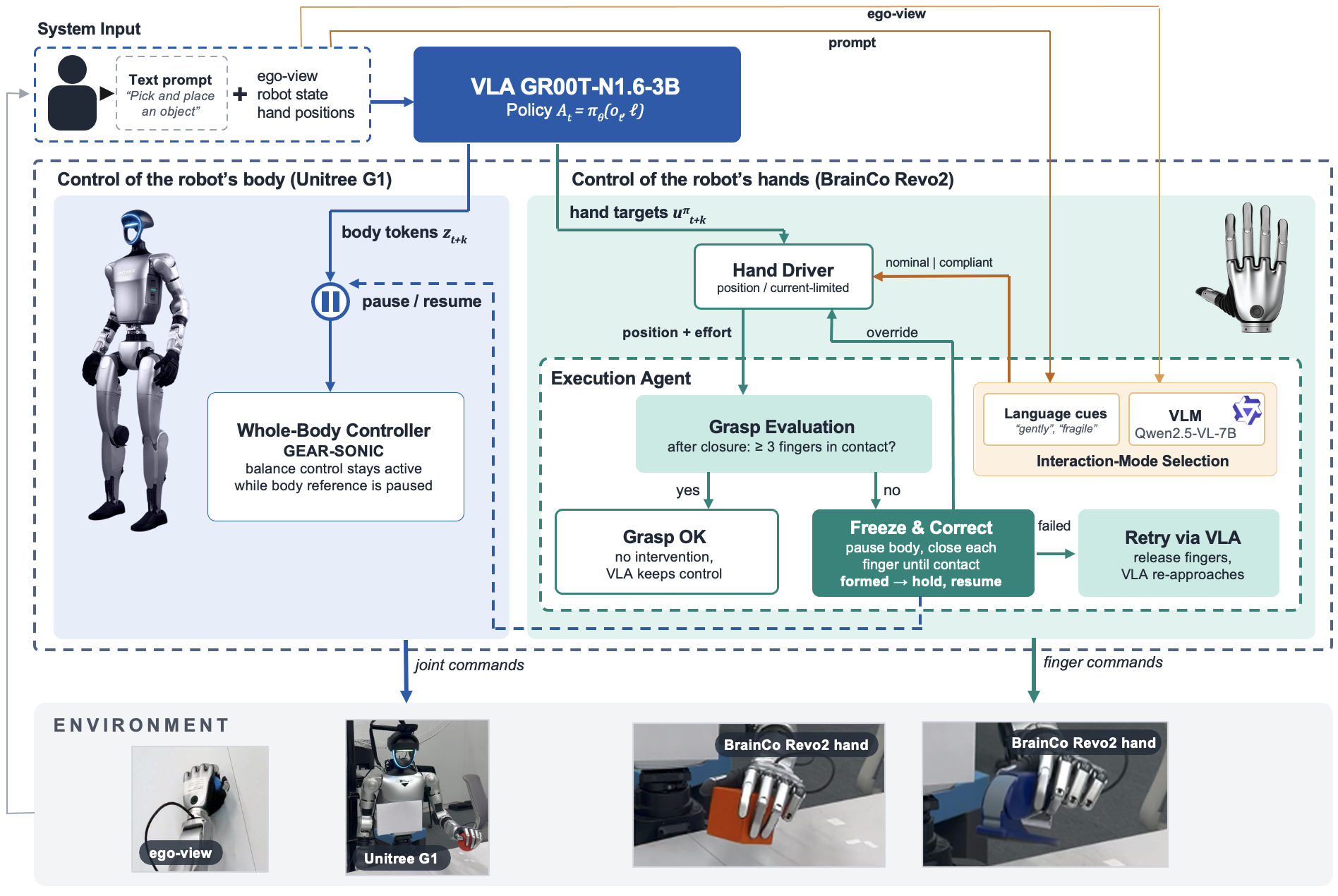}
\caption{System architecture of AgenticTactileVLA. The fixed VLA provides the
nominal manipulation strategy, while the execution supervisor monitors its
physical realization and intervenes only when adaptation is required.}
\vspace{-3mm}
\label{fig:architecture}
\end{figure*}

The architecture separates policy prediction from physical execution.
The fixed VLA generates body-motion tokens and hand targets, while the
execution supervisor monitors how these commands are physically realized.
When the policy-generated grasp is sufficiently realized, execution remains
unchanged. Otherwise, the supervisor can pause body-reference progression,
adapt individual fingers, or return control to the VLA for another approach.
A separate interaction-mode layer selects nominal or compliant hand control
before closure, while whole-body stabilization remains active throughout
execution (Fig.~\ref{fig:architecture}).

\subsection{System Overview and Execution Supervisor}
\label{sec:supervisor}

\paragraph{Nominal VLA execution}
GR00T-N1.6-3B \cite{groot}, fine-tuned on teleoperated manipulation
demonstrations, is deployed on a Unitree G1 with a BrainCo Revo2 hand and
remains fixed during agent-assisted execution.

Let $o_t$ denote the policy observation, including the ego-view image, robot
state, and measured hand positions, and let $\ell$ denote the natural-language
instruction. The VLA produces an action chunk
\begin{equation}
A_t=\pi_\theta(o_t,\ell)
=
\left(
z_{t+k},\mathbf{u}^{\pi}_{t+k}
\right)_{k=0}^{H-1},
\label{eq:policy}
\end{equation}
where $z_{t+k}$ is the body-motion token and
$\mathbf{u}^{\pi}_{t+k}\in[0,1]^6$ contains the nominal position targets for
the six hand channels.

GEAR-SONIC \cite{sonic} realizes the body-motion tokens, while a dedicated
driver executes thumb flexion (0), thumb opposition (1), index (2), middle
(3), ring (4), and little finger (5). Thumb opposition preserves the
policy-selected grasp arrangement; execution-time correction modifies only
the flexion channels.

The hand publishes measured finger positions
$\mathbf{q}_t\in[0,1]^6$ and a motor-effort proxy $\mathbf{e}_t$. These signals
provide the proprioceptive feedback used by the supervisor. Motor effort is
treated as a device-specific indicator of interaction resistance, not as a
calibrated measurement of contact force.

\paragraph{Agentic execution supervisor}
The execution agent wraps the fixed VLA as a stateful supervisory layer:
\begin{equation}
\left(
m_t,\gamma_t,\mathbf{r}_t,b_t
\right)
=
\mathcal{S}
\left(
A_t,\mathcal{H}_t,\ell,I_t,\sigma_t
\right),
\label{eq:supervisor}
\end{equation}
where $\mathcal{H}_t$ stores issued hand references, measured positions, and
motor-effort feedback; $I_t$ is the visual observation and $\sigma_t$ the
internal execution state.

The outputs define execution authority. The mode $m_t$ selects the hand-control
regime, $\gamma_t$ selects nominal execution or intervention,
$\mathbf{r}_t$ contains override finger references, and $b_t$ gates
progression of the body reference. The agent does not replace the VLA arm
policy: body motion continues to originate from policy-generated tokens.

The supervisor operates at the boundary between \emph{policy completion} and
\emph{physical grasp completion}. A closure is visible in the action chunk as
progressive flexion-target motion. In the reference realization, the closure
is armed when any flexion target advances by $0.015$ over $0.25$\,s and
reaches its terminal phase when every flexion target varies by less than
$0.005$ over $0.12$\,s. This terminal phase defines the grasp-assessment point:
the VLA has completed its commanded grasp, but the supervisor must still
determine whether sufficient physical support has actually been established.

Position-based support evidence is accumulated from arming, while
effort-relative evidence is introduced only after entering \state{Adjust}.
If support is sufficient, the supervisor remains transparent. Otherwise,
body-reference progression is paused and the flexion targets are locally
corrected. Missing evidence at this stage does not by itself prove absence of
physical contact.

An accepted correction is retained while body-reference progression resumes.
A rejected correction releases the hand override and returns execution
authority to the VLA. At authority transitions, stale queued chunks are
discarded and the next VLA chunk is generated from a fresh observation;
whole-body stabilization holds the current body reference while the new chunk
is pending. Thus, the corrected physical hand state can directly condition
subsequent VLA execution.

\paragraph{Control branches and execution timing}
The supervisor has two physical execution branches. The position branch
preserves the VLA grasp strategy and supports contact-guided finger correction.
The compliant branch uses current-limited interaction control and can be
selected before closure when language or visual context indicates that
interaction effort should be constrained.

Finger references are issued at 50\,Hz. When $b_t=1$, only progression of the
body reference is paused; whole-body stabilization remains active. Table~\ref{tab:states} summarizes the supervisory states and authority
transitions.

\begin{table}[!t]
\caption{Execution-agent states. $b_t$ gates body-reference progression; the
override column indicates whether the agent replaces the nominal hand target.}
\label{tab:states}
\centering\small
\setlength{\tabcolsep}{4pt}
\begin{tabularx}{\columnwidth}{lccY}
\toprule
State & $b_t$ & Override & Transition / exit condition\\
\midrule
\state{Idle}
& 0 & no
& upcoming closure in the action chunk\\
\state{Armed}
& 0 & no
& terminal grasp assessment\\
\state{Adjust}
& 1 & yes
& grasp verdict or timeout\\
\state{Hold}
& 0 & yes
& policy reopens the hand\\
\bottomrule
\end{tabularx}
\end{table}

\subsection{Selective Intervention}
\label{sec:intervention}

Selective intervention determines when physical execution should transfer from
the VLA to the supervisor. It is evaluated only at the grasp-assessment point;
before this point, missing contact is expected and does not trigger correction.

Let $\hat{c}_i(t)\in\{0,1\}$ denote the position-support evidence accumulated
since arming for the monitored flexion channels
\[
\mathcal{F}=\{0,2,3,4,5\},
\]
and let $g_t$ indicate that the VLA has reached the terminal phase of its
commanded grasp. Intervention is defined as
\begin{equation}
\eta_t =
g_t\,\mathbb{I}\!\left[
\sum_{i\in\mathcal{F}}\hat{c}_i(t)<3
\right],
\label{eq:intervention}
\end{equation}
where $\eta_t=1$ transfers hand execution from the nominal VLA command to the
execution agent.

At least three support latches allow nominal execution to continue. Otherwise,
the supervisor sets $b_t=1$, overrides the flexion targets, and enters
\state{Adjust}. The three-support criterion is an operational acceptance rule,
not a guarantee of grasp stability.

\subsection{Proprioceptive Contact Estimation}
\label{sec:contact}

Contact is estimated only from proprioceptive signals available from the hand.
For each monitored flexion channel $i\in\mathcal{F}$, let $u_i(t)$ denote
commanded position, $q_i(t)\in[0,1]$ measured normalized position, and $e_i(t)$
device-specific motor effort. Thumb opposition remains at the policy target.

The main cue is the mismatch between commanded and measured finger motion.
Over a history window $W$,
\begin{align}
\Delta u_i(t) &= u_i(t)-u_i(t-W),\\
\Delta q_i(t) &= q_i(t)-q_i(t-W).
\label{eq:progress}
\end{align}
When the command continues to advance but measured motion is substantially
smaller, the finger is interpreted as encountering resistance.

Let $E_i(t)\in\{0,1\}$ denote detector eligibility. Eligibility requires a
complete history window without reopening and at least $\delta_u$ commanded
advance; a saturated target does not create additional evidence. Position-based
stall is
\begin{equation}
s_i(t)=E_i(t)\,\mathbb{I}\!\left[
\Delta u_i(t)\geq\delta_u
\land
\Delta q_i(t)<\rho_s\Delta u_i(t)
\right],
\label{eq:stall}
\end{equation}
where $\rho_s$ bounds the fraction of commanded motion reproduced by the
measured finger.

Motor effort provides complementary evidence during adjustment. At entry to
\state{Adjust}, the current feedback sample establishes a fixed per-channel
baseline $b_i$, and the effort-duration accumulator is reset. Let $D_i(t)$ be
the continuous duration for which
\[
|e_i(t)-b_i|>\delta_e .
\]
Sustained effort deviation is combined with incomplete tracking:
\begin{equation}
\begin{split}
d_i(t)=E_i(t)\,\mathbb{I}\big[
&D_i(t)\geq T_e
\land \Delta u_i(t)\geq\delta_u\\
&{}\land \Delta q_i(t)<\rho_d\Delta u_i(t)
\big],
\end{split}
\label{eq:duration}
\end{equation}
where $T_e$ is the required effort-deviation duration and $\rho_d$ the
corresponding tracking-ratio threshold.

Support for finger $i$ is latched when $s_i(t)\lor d_i(t)$ activates.
Pre-assessment position latches are preserved, effort-assisted detections
accumulate only in \state{Adjust}, and reopening resets all latches. Missing
activation can therefore indicate compliance or insufficient excitation rather
than true absence of contact.

The deployed detector parameters are
\[
W=0.25\,\mathrm{s},\qquad
\delta_u=0.015,\qquad
\rho_s=0.3,
\]
\[
\rho_d=0.7,\qquad
\delta_e=0.3,\qquad
T_e=0.06\,\mathrm{s}.
\]

\paragraph{Parameter selection}
Detector thresholds were selected from 40 calibration closures on the five
training objects (eight per object), separate from all evaluation trials.
Calibration targeted stable finger stopping and suppression of short effort
transients. Held-out objects were not used for tuning, and all thresholds,
preload, timing parameters, and controller gains were frozen before
evaluation.

Highly compliant objects remain challenging because deformation can preserve
command tracking despite physical contact, reducing the command--response
mismatch required for detection.

\subsection{Object-Adaptive Finger Correction and Acceptance}
\label{sec:correction}

Once intervention transfers execution to the supervisor, each unsupported
finger advances independently toward its permitted travel limit. When support
is detected, the reference is fixed at the measured position plus a normalized
preload of $0.02$, clipped to the travel limit; further advancement of that
finger stops. Correction starts from measured positions and advances in
increments of $0.004$ every 20\,ms.

Independent stopping transforms the VLA-proposed terminal hand configuration
into object-dependent finger references: different fingers may stop at
different depths and times, while the policy-selected thumb-opposition target
preserves the overall grasp arrangement.

Reaching the travel limit without detector activation does not establish
contact, because insufficient remaining motion may prevent the history and
progress requirements of \eqref{eq:stall} from being satisfied.

Adjustment lasts at least 0.6\,s and at most 4.0\,s. After the minimum
duration, it terminates when all monitored fingers have either latched support
or exhausted their permitted travel, or when the timeout is reached. The final
support count defines
\begin{equation}
v=
\begin{cases}
\formed,
& \displaystyle\sum_{i\in\mathcal{F}} c_i \geq 3,\\
\emptygrasp,
& \displaystyle\sum_{i\in\mathcal{F}} c_i = 0,\\
\crooked,
& \text{otherwise}.
\end{cases}
\label{eq:verdict}
\end{equation}

The verdict depends only on total support. \crooked\ denotes insufficient
support evidence for acceptance rather than geometric misalignment, while
\emptygrasp\ denotes no detected support.

A \formed\ verdict retains the corrected configuration and releases the body
gate. An \emptygrasp\ or \crooked\ verdict releases the hand override, allowing
subsequent VLA actions to reopen the hand and generate another approach.
Releasing the override does not itself command reopening.

To prevent repeated intervention on one closure, re-entry is blocked until
the hand has reopened. A decrease of at least $0.05$ in policy-commanded mean
flexion relative to terminal assessment, sustained for $0.12$\,s, first
releases a retained override. Rearming then requires both commanded and
measured mean flexion to remain at least $0.05$ below their terminal-assessment
values for $0.12$\,s. Each closure therefore produces at most one correction
episode, and all pauses and retries consume the same trial budget.

\subsection{Language- and Vision-Conditioned Mode Selection}
\label{sec:mode_selection}

Interaction-mode selection chooses between nominal position control and
compliant control. Explicit handling cues such as ``gently'', ``carefully'',
or ``fragile'' activate compliant mode.

When no explicit cue is present, Qwen2.5-VL-7B-Instruct \cite{qwen} selects
between the two predefined modes from image and task context. The VLM runs
asynchronously while the hand is open, with one decision cached per grasp
episode. Language cues take priority; missing, late, or malformed VLM
responses default to nominal control.

The VLM selects only the interaction mode and does not estimate material
properties or generate controller parameters. Selection is performed before
closure because a fragile object may deform before sufficient proprioceptive
resistance can be detected.

To isolate mode selection in Section~\ref{sec:cup}, the VLA receives the same
task instruction in every condition. Handling cues are provided only to the
interaction parser, while the VLM receives a fixed image-and-task prompt
without those cues.

\subsection{Compliant Control via Current-Limited Interaction}
\label{sec:compliant}

For compliant interaction, the agent activates a custom current-limited
controller:
\begin{equation}
I_i^{\mathrm{cmd}}(t)=
\operatorname{sat}_{[-I_{\max},I_{\max}]}
\left(
K_p[q_i^d(t)-q_i(t)]
-
K_d\bar{\dot q}_i(t)
\right),
\end{equation}
where $q_i^d$ and $q_i$ are normalized hand-interface positions,
$\bar{\dot q}_i$ is the exponentially smoothed finger velocity, and
$I_i^{\mathrm{cmd}}$ is a device-level actuator command rather than calibrated
motor current.

We use $K_p=5.0$ and $K_d=0.05$. These are controller-interface gains rather
than physical stiffness or damping coefficients. The controller runs every
20\,ms and bounds the actuator command by $I_{\max}$ while driving the fingers
toward a predefined grasp configuration.

Raw command limits of 150 and 200 device units correspond to normalized
controller-interface limits of $I_{\max}=0.15$ and $0.20$, respectively.
These quantities do not represent calibrated motor current or fingertip force.

Compliant mode is activated before closure to prevent the nominal
position-control path from applying an initial high-effort squeeze. The current
implementation supports compliant closure and holding but not autonomous
transport; fragile-object experiments therefore evaluate object integrity
during closure and holding only.

\section{Experimental Results}
\label{sec:results}

We evaluate AgenticTactileVLA through a sequence of controlled experiments.
The evaluation first establishes the task-level benefit of execution-time
supervision, then separates this effect from unconditional finger closure.
We further examine sensitivity to initial-pose perturbations, the physical
meaning and failure modes of proprioceptive verification, and contextual
selection of the interaction regime for fragile objects.

\subsection{Experimental Protocol}
\label{sec:protocol}

All experiments use the same fixed VLA checkpoint. The five \emph{held-out}
test objects were excluded from VLA training, detector calibration, and
controller development.

A \emph{trial} starts from an initialized scene and may contain multiple
closures. A \emph{retry} requires policy-commanded reopening followed by a new
closure. Task completion is recorded independently of the supervisor.

The primary comparison uses randomized, interleaved matched blocks under a
shared execution budget. Within each block, all conditions start from a common
initial configuration. A trial terminates at successful placement, 60\,s, or
three arm approaches, whichever occurs first. No started trial is excluded.

Paired task-level comparisons use two-sided exact McNemar tests with Holm
correction over all pairwise contrasts reported within each experiment.

\subsection{Main Comparison}
\label{sec:controlled}

The primary experiment compares the base VLA, unconditional close-to-stall
control, and the full position-branch supervisor over 75 matched blocks
(15 per held-out object).

The close-to-stall baseline isolates a simpler explanation for improvement:
the robot may benefit merely from allowing the fingers to close farther. At
each terminal grasp, this baseline pauses body-reference progression and
advances finger flexion until the position-stall condition in
\eqref{eq:stall}, the travel limit, or the shared 4.0\,s timeout. Unlike the
supervisor, it acts at every closure and has no contact-count verdict or
verdict-dependent release.

\begin{table}[H]
\vspace{-4mm}
\caption{Fixed-budget task completion under randomized, interleaved conditions.
All conditions use 15 trials per held-out object; intervals are 95\% Wilson.}
\label{tab:controlled}
\centering\small
\setlength{\tabcolsep}{3pt}
\begin{tabularx}{\columnwidth}{Yccc}
\toprule
Object & VLA & VLA + stall & Supervisor\\
\midrule
Cube             & 9/15  & 11/15 & 13/15\\
Fish             & 9/15  & 11/15 & 13/15\\
Soy sauce bottle & 11/15 & 12/15 & 14/15\\
Thermos lid      & 7/15  & 8/15  & 10/15\\
Hand cream tube  & 10/15 & 12/15 & 13/15\\
\midrule
Pooled            & 46/75 & 54/75 & 63/75\\
Rate              & 61.3\% & 72.0\% & 84.0\%\\
95\% CI (\%)      & 50.0--71.5 & 61.0--80.9 & 74.1--90.6\\
\bottomrule
\end{tabularx}
\end{table}

Completion increases from 61.3\% for the base VLA and 72.0\% for
unconditional close-to-stall control to 84.0\% with the supervisor. In the
matched blocks, the supervisor succeeds where the base VLA fails in 20 cases,
versus three in the opposite direction, yielding $p=0.001$ after Holm
correction. Against unconditional closure, the corresponding discordance is
11/2 ($p=0.045$, Holm-corrected) (Table \ref{tab:controlled}).

Unconditional closure alone exceeds the base VLA in 12/4 discordant blocks
($p=0.077$), recovering roughly half of the supervisor's gain but not
significantly outperforming the base policy. This indicates that additional
closure contributes to performance, but does not fully account for the
supervisor's gain.

The improvement is also consistent across all five objects: the supervisor
gains 3--4 successful trials over the base VLA and 1--2 over unconditional
closure for every object. Thus, the primary result is not driven by one
particularly favorable test object.

The supervisor intervenes in 27/75 trials (36\%), producing 36 correction
episodes, whereas the unconditional baseline extends finger closure at every
terminal grasp, resulting in 101 extension episodes across the same 75 trials.
This motivates the component ablation below, which separates structured
supervisory decisions from simply extending every closure.

\subsection{Component Ablation}
\label{sec:ablations}

To determine which components of the execution supervisor drive the observed
improvement, we conduct a component ablation over 100 matched blocks. The
analysis examines whether the gain arises simply from allowing the fingers to
close farther, whether selective intervention reduces unnecessary corrections,
and whether effort evidence and verdict-dependent release provide additional
benefit.

We evaluate five variants. Here, \emph{U} denotes unconditional intervention
at every closure, while \emph{S} denotes selective intervention only when the
supervisor detects insufficient support. \emph{U-stall} simply extends every
closure until position-based stall and serves as the additional-closure
baseline. \emph{U-full} applies the complete contact detector and grasp-verdict
logic at every closure, testing whether structured supervision provides benefit
even without selectivity. \emph{S-position} retains selective intervention but
removes motor-effort evidence. \emph{S-retain} retains selective intervention
but disables verdict-dependent release and policy-mediated retry.
\emph{S-full} is the complete selective supervisor
(Table \ref{tab:ablations}).

\begin{table}[H]
\caption{Component-controlled ablation over 100 matched blocks. Corrections
count intervention episodes.}
\label{tab:ablations}
\centering\small
\setlength{\tabcolsep}{4pt}
\begin{tabularx}{\columnwidth}{Yccc}
\toprule
Variant & Completion & Rate & Corrections\\
\midrule
U-stall     & 72/100 & 72\% & 135\\
U-full      & 84/100 & 84\% & 134\\
S-position  & 79/100 & 79\% & 49\\
S-retain    & 77/100 & 77\% & 47\\
S-full      & 86/100 & 86\% & 46\\
\bottomrule
\end{tabularx}
\end{table}

The full supervisor reaches 86\% completion versus 72\% for position-only
stall control (Holm-corrected $p=0.047$). The always-on variants show the same
direction of effect: U-full reaches 84\% compared with 72\% for U-stall,
although this contrast is not statistically resolved after correction
($p=0.091$). Together, these results indicate that the gain is not explained
by extended finger closure alone.

Comparing U-full with S-full isolates the role of selectivity. Their completion
rates are similar (84\% and 86\%), but selective supervision reduces the
number of correction episodes from 134 to 46, a 65.7\% reduction. No
completion difference is detected between the two variants
($p=0.804$; paired 95\% CI $-5.8$ to $+9.8$ points). Thus, selectivity mainly
reduces unnecessary intervention while preserving the observed task-level
performance.

Finally, S-position and S-retain test the contribution of individual
supervisory mechanisms. Removing effort evidence reduces completion to 79\%,
while disabling verdict-dependent release reduces it to 77\%. Neither
individual contrast is statistically resolved after correction
($p=0.378$ and $0.235$, respectively). The ablation therefore supports the
complete supervisory design over simple position-only closure, while the
independent contribution of these two components remains unresolved.

\subsection{Pose-Perturbation Evaluation}
\label{sec:pose_stress}

The previous experiments use matched initial configurations. We therefore test
whether the supervisor's benefit disappears when the initial object pose is
perturbed. The same five objects are evaluated under nominal poses, horizontal
translations within 20\,mm, and yaw changes within $15^\circ$, with 50 matched
trials per regime and condition.


Performance decreases under perturbation for all methods, but the supervisor
remains the strongest condition in every regime. Its gain over the base VLA
is 22, 28, and 28 percentage points for nominal, translated, and rotated
configurations, respectively. Pooled over regimes, the supervisor exceeds
unconditional closure in 26/6 discordant blocks ($p<0.001$); the individual
per-regime contrasts at 50 trials are underpowered (Table \ref{tab:pose_stress}).

\begin{table}[H]
\caption{Task completion under initial-pose perturbations.}
\label{tab:pose_stress}
\centering\small
\setlength{\tabcolsep}{3pt}
\begin{tabularx}{\columnwidth}{Yccc}
\toprule
Regime & VLA & VLA + stall & Supervisor\\
\midrule
Nominal      & 32/50 & 37/50 & 43/50\\
Translation  & 25/50 & 32/50 & 39/50\\
Yaw          & 23/50 & 30/50 & 37/50\\
\midrule
Pooled       & 80/150 & 99/150 & 119/150\\
\bottomrule
\end{tabularx}
\end{table}

The important observation is therefore not pose invariance, but persistence
of the execution-time benefit within the tested perturbation range. The
supervisor still depends on a feasible policy-generated approach and is not
designed to recover from arbitrary approach errors.

\subsection{Proprioceptive Verification and Failure Modes}
\label{sec:predicate}

The previous results show that the supervisor improves task completion, but
they do not establish whether its internal support decisions correspond to a
meaningful physical state. We therefore examine the contact-support mechanism
separately from task success.

Across 70 diagnostic trials, the supervisor produces 85 correction events:
46 \formed\ verdicts, 26 \crooked\ verdicts, and 13 \emptygrasp\ verdicts.
These internal decisions are not treated as contact-detection accuracy because
independent per-finger contact labels are unavailable.

\begin{figure}[H]
\centering
\includegraphics[
   width=\columnwidth,
   trim=12mm 7mm -3mm 7mm,
   clip
]{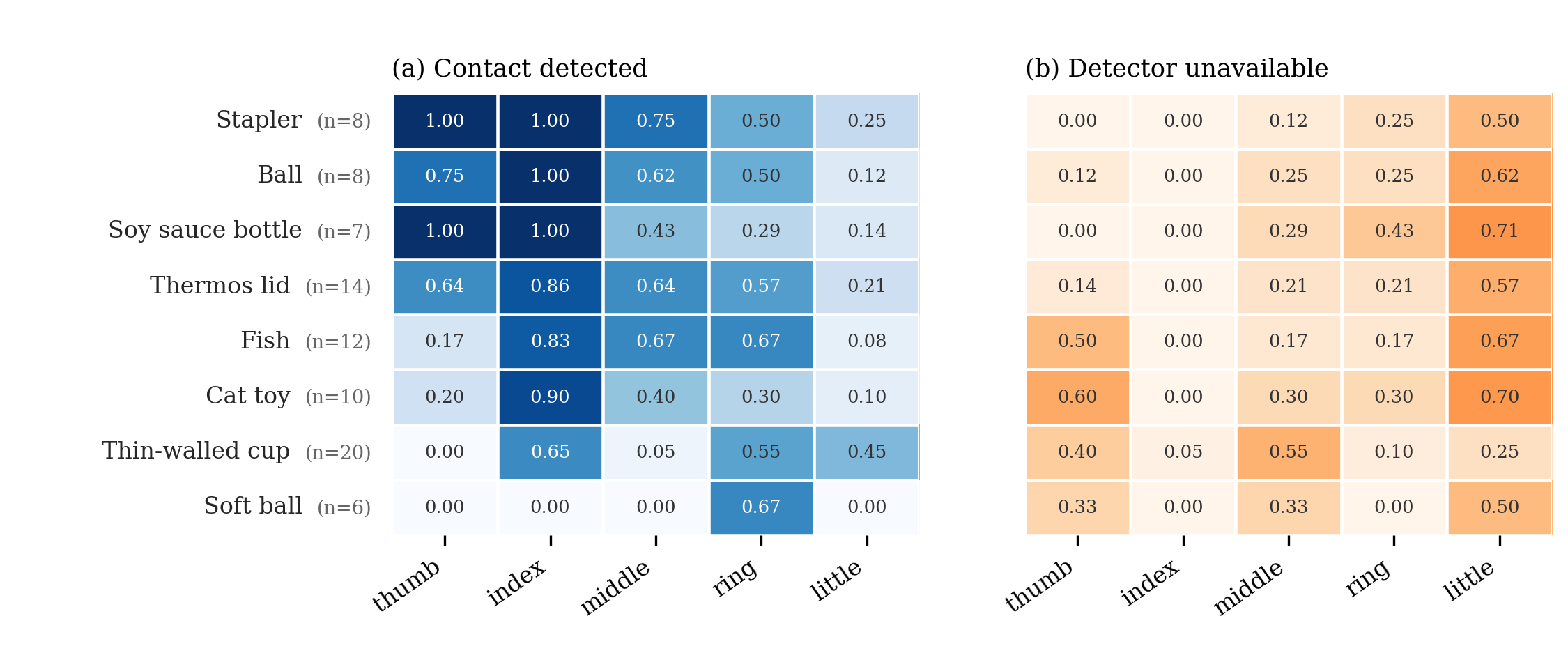}
\caption{Diagnostic contact detection (a) and detector ineligibility due to
insufficient remaining finger travel (b).}
\label{fig:contactmap}
\vspace{-3mm}
\end{figure}

Contact patterns vary by object (Fig.~\ref{fig:contactmap}). In particular,
highly compliant objects expose a characteristic limitation of proprioceptive
contact inference.

\paragraph{Compliance-related false rejection}
The stapler produces a \formed\ verdict in 8/8 correction events and the ball
in 5/8, whereas the soft ball produces none in six events. Video confirms that
the soft ball is physically enclosed in all six cases. Its deformation allows
the fingers to continue following their commanded motion, reducing the
command--response mismatch used as contact evidence. Four events are therefore
classified as \crooked\ and two as \emptygrasp.

Despite these false rejections, five of the six trials eventually complete:
three retain the object through placement and two succeed after a
policy-generated retry. This shows that incorrect internal verdicts do not guarantee task failure, but highlights a clear limitation of motion-based proprioceptive inference.

\paragraph{External retention audit}
To determine whether the support predicate nevertheless predicts a useful
physical outcome, we perform 120 independent standardized closures (Table \ref{tab:retention_audit}). The grasp
configuration is retained for a 5\,cm diagnostic lift followed by a 2\,s hold,
providing an external retention label independent of the supervisor's verdict.

\begin{table}[t]
\caption{Support verdict versus object retention during the diagnostic lift.}
\label{tab:retention_audit}
\centering\small
\setlength{\tabcolsep}{3pt}
\begin{tabularx}{\columnwidth}{Yccc}
\toprule
Support predicate & Retained & Not retained & Total\\
\midrule
At least three & 70 & 6  & 76\\
Below three    & 18 & 26 & 44\\
\midrule
Total          & 88 & 32 & 120\\
\bottomrule
\end{tabularx}
\vspace{-6mm}
\end{table}

Accepted grasps retain the object in 70/76 cases (92.1\%), showing that a
positive support verdict is strongly associated with physical retention.
Rejection is less reliable: 18/44 rejected configurations would still retain
the object, corresponding to an NPV of 59.1\%.

This asymmetry becomes stronger on held-out objects. There, retention given
acceptance is 88.9\%, while 25.6\% of retainable configurations are falsely
rejected. The support rule is therefore useful but conservative: acceptance
provides strong evidence of a physically useful grasp, whereas rejection is
less reliable, particularly for compliant or previously unseen objects.

\subsection{Contextual Interaction-Mode Selection}
\label{sec:cup}

The previous experiments adapt grasp realization during or after contact.
Fragile objects pose a different challenge: damage may occur before sufficient
resistance is observed. We therefore test whether the agent can select an
alternative interaction regime before closure.

A thin-walled cup is evaluated under position control, instruction-based
routing, VLM-based routing, combined instruction--VLM routing, and
always-compliant control (Table \ref{tab:modes}).

\begin{figure}[H]
\centering
\includegraphics[width=0.6\columnwidth]{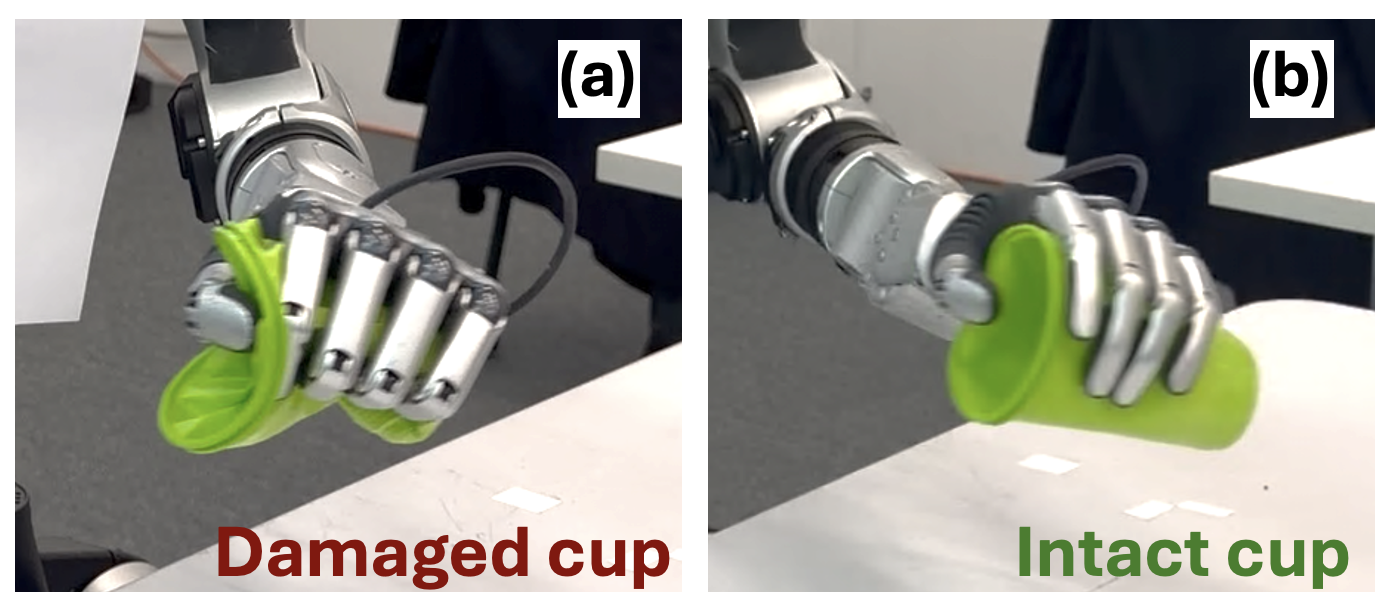}
\caption{Representative thin-walled-cup outcomes.
(a) Damage under nominal position control.
(b) Preserved integrity under compliant control.}
\label{fig:cup_outcomes}
\vspace{-2.5mm}
\end{figure}

Nominal position control damages all five cups, whereas instruction-based and
combined routing preserve 5/5. VLM-only routing preserves 4/5; in the remaining
trial, the correct compliant mode is selected after closure begins
(Fig.~\ref{fig:cup_outcomes}(a)), showing that correct semantic selection must
also arrive in time to affect the interaction.

\begin{table}[H]
\caption{Thin-walled-cup integrity and compliant-mode activation.}
\label{tab:modes}
\centering\small
\setlength{\tabcolsep}{3pt}
\begin{tabularx}{\columnwidth}{Ycccc}
\toprule
Mode selected by & Trials & Intact & Damaged & Compliant active\\
\midrule
No selection        & 5 & 0 & 5 & 0/5\\
Instruction cue     & 5 & 5 & 0 & 5/5\\
VLM from image      & 5 & 4 & 1 & 4/5\\
Instruction + VLM   & 5 & 5 & 0 & 5/5\\
Always compliant    & 5 & 5 & 0 & 5/5\\
\bottomrule
\end{tabularx}
\vspace{-4mm}
\end{table}


Always-compliant control also preserves 5/5 cups, so the experiment
demonstrates successful contextual routing rather than superiority over
always-compliant execution (Fig.~\ref{fig:cup_outcomes}(b)). Since the cup is a
training object, this does not establish held-out fragile-object generalization.

On three additional rigid objects, the intended nominal mode is selected in
14/15 trials. This confirms nonconstant routing between the two predefined
interaction regimes, but not a general material classifier.

\section{Discussion and Limitations}
\label{sec:limitations}

The experiments show that a fixed VLA benefits from execution-time adaptation.
The supervisor improves completion over both the fixed VLA and unconditional
close-to-stall control, with gains on every held-out object. Additional closure
explains only part of the improvement, while the ablation supports structured
supervision over position-only stall control.

Selective intervention reduces correction episodes by 65.7\% relative to the
always-on controller without detected completion loss, and the benefit persists
under moderate pose variation.

The retention audit shows that accepted configurations retain the object in
70/76 cases (92.1\% pooled; 88.9\% on held-out objects), while 18/44 rejected
configurations would still retain it. The support predicate is useful but
conservative, especially for compliant objects. False rejections can still be
recovered through policy-mediated retry.

Interaction-mode selection extends adaptation before contact. Instruction-based
and combined routing preserve 5/5 fragile cups, while VLM-only routing preserves
4/5 due to late mode selection. On rigid objects, the intended nominal mode is
selected in 14/15 trials.

Limitations remain. The ablation does not isolate the contributions of effort
evidence and verdict-dependent release, and the lack of a detected difference
between selective and always-on supervision does not imply equivalence. The
supervisor relies on proprioceptive signals rather than calibrated force and
cannot recover from arbitrary approach errors. The fragile-object study uses a
training object, and compliant control is evaluated only for static grasping.
Validation requires unseen objects and diverse embodiments.

Overall, the results support execution-time adaptation as a complement to a
reusable fixed VLA.

\balance
\section{Conclusion}

AgenticTactileVLA introduces contact-guided execution-time supervision for a
fixed VLA. Rather than retraining for each object, it preserves the predicted
strategy while adapting its physical realization through selective
intervention, per-finger correction, grasp acceptance or policy-mediated retry,
and interaction-mode selection.

On five held-out objects, the supervisor achieves 84.0\% task completion versus
61.3\% for the base VLA and 72.0\% for unconditional close-to-stall control.
The benefit persists under pose perturbations, while selective supervision
reduces interventions and accepted configurations provide support.
Fragile-object experiments extend adaptation to pre-contact mode selection.

These results support execution-time adaptation toward more scalable
object-level generalization of fixed VLA policies without object-specific
retraining or additional tactile hardware.
\vspace{-1mm}

\bibliographystyle{IEEEtran}
\bibliography{IEEEexample}

\end{document}